\documentclass{bmvc2k}

\usepackage{epsfig}
\usepackage{graphicx}
\usepackage{amsmath}
\usepackage{amssymb}

\usepackage{booktabs}

\usepackage{multirow}
\usepackage{multicol}
\usepackage{cuted}
\usepackage{capt-of}

\newcommand{\nameMethod}{MIGS}

\def\thefootnote{*}\footnotetext{The first two authors contributed equally to this work.}

\title{\hspace{20pt}MIGS: Meta Image Generation from Scene \\\hspace{15pt} Graphs}

\addauthor{Azade Farshad\thefootnote{}}{azade.farshad@tum.de}{1} %
\addauthor{Sabrina Musatian\thefootnote{}}{sabrina.musatian@tum.de}{1}
\addauthor{Helisa Dhamo}{helisa.dhamo@tum.de}{1}
\addauthor{Nassir Navab}{nassir.navab@tum.de}{1}

\addinstitution{
Technical University of Munich\\
 Munich, Germany
}

\runninghead{MIGS}{Meta Image Generation from Scene Graphs}

\def\etal{\emph{et al}\bmvaOneDot}
\def\ie{\emph{i.e}\bmvaOneDot} 
\begin{document}

\maketitle
\begin{center}
\centering
\vspace{-0.6cm}
    \includegraphics[width=0.88\textwidth]{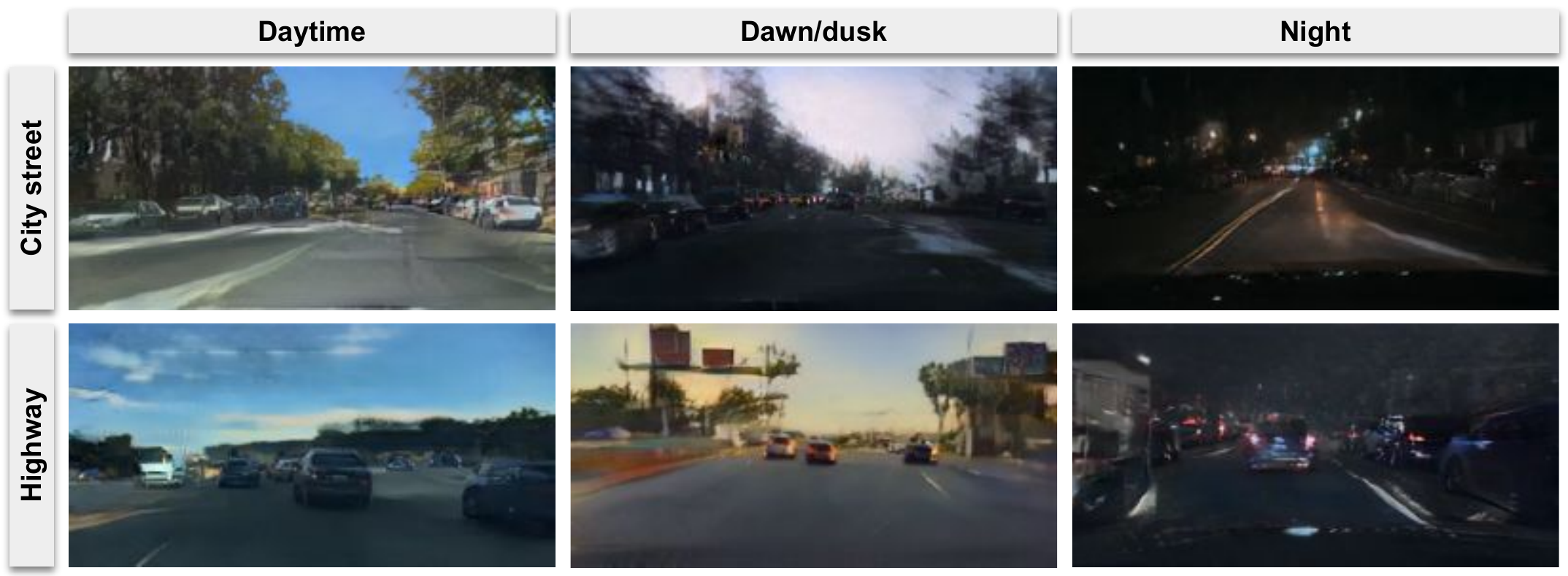}
    \captionof{figure}{Sample of images generated by MIGS over a variety of scene attributes from BDD dataset~\cite{seita2018bdd100k}. The model is trained on 10-shot data for generating the images.}
    \label{fig:teaser}
\end{center}

\begin{abstract}
Generation of images from scene graphs is a promising direction towards explicit scene generation and manipulation. However, the images generated from the scene graphs lack quality, which in part comes due to high difficulty and diversity in the data. We propose MIGS (Meta Image Generation from Scene Graphs), a meta-learning based approach for few-shot image generation from graphs that enables adapting the model to different scenes and increases the image quality by training on diverse sets of tasks. By sampling the data in a task-driven fashion, we train the generator using meta-learning on different sets of tasks that are categorized based on the scene attributes. Our results show that using this meta-learning approach for the generation of images from scene graphs achieves state-of-the-art performance in terms of image quality and capturing the semantic relationships in the scene. Project Website: \url{https://migs2021.github.io/}
\end{abstract}

\section{Introduction}
The task of high-quality image generation and manipulation has been capturing the attention of researchers for many years. Recent advances in deep learning for unconditional image synthesis~\cite{karras2018Progr, karras2019style, karras2020analyzing, karras2020Ada} has led to producing high-quality images that are often indistingu-
\newpage
\noindent ishable from the real ones by a human. Some impressive results have been also achieved for the class-conditional image generation~\cite{bigGan} and generation from semantic segmentation masks~\cite{park2019semantic}. While the latter method allows for pixel-level control on the image, it requires a segmentation map which may be practically hard to obtain. An easier alternative to represent image semantics is a scene graph~\cite{johnson2015image}, \ie a graph where nodes correspond to the objects and edges define the relationships between them. Image generation from such graphs is attractive as they allow full control over the semantics and it is easy to modify them in case of scene editing. Image generation from scene graphs has been introduced in~\cite{johnson2018image}. While the results are encouraging, the images generated by this model lack quality when trained on datasets with diverse scenes, as the network struggles to learn meaningful representations to accommodate such discrepancies. To counterpart this problem we suggest using meta-learning in order to help the network focus its attention on specific tasks during training. Yet such a model is able to quickly adjust to a wide set of tasks during testing with only a few training samples available. Additionally, our approach allows us to introduce the task of few-shot learning for scene graph to image generation problem, which is an attractive scenario as it can help to generate semantically meaningful images in applications with a scarce amount of data. The contributions of our work can be summarized as follows:
\begin{itemize}
    \item A novel meta-learning approach for the generation of images from scene graphs, which achieves state-of-the-art results compared to previous work
    \item Introduction of the few-shot learning problem for scene graph to image generation
    \item A novel task sampling method for scenes in the wild that can benefit other image generation with meta-learning scenarios.
\end{itemize}

We evaluate our proposed method on automatically generated scene graphs for Berkeley Deep Drive~\cite{seita2018bdd100k}, Action Genome~\cite{ji2020action} and Visual Genome~\cite{krishna2017visual} datasets, showing superior results compared to the baselines, qualitatively and quantitatively. This is also verified by performing a user study on the quality of images. The source code of this work is provided in the supplement and will be publicly released upon its acceptance.
\section{Related Work}

\paragraph{Image generation}

Recent advances in generative models, in particular, Generative Adversarial Networks (GANs)~\cite{goodfellow2014generative} have boosted the quality of image generation. A line of works explore generative models for unconditional image generation ~\cite{karras2018Progr, karras2019style, karras2020analyzing, karras2020Ada}. Image generation models have also been explored conditionally, with a diverse set of priors such as semantic segmentation maps~\cite{chen2017photographic,wang2018high,park2019semantic}, natural language descriptions~\cite{reed2016generative,zhang2018photographic,zhang2017stackgan,li2018storygan} or translating from one image domain to another using paired~\cite{isola2017image} or unpaired data~\cite{zhu2017unpaired}. 
Most related to our approach are methods that generate images from scene graphs~\cite{johnson2018image}.

\paragraph{Image generation from scene graphs}

Scene graphs~\cite{johnson2015image} refer to representations that describes images, where nodes are objects, and the edges represent relationships between them. With the recent rise of large-scale scene graph datasets, such as Visual Genome~\cite{krishna2017visual} a diverse set of scene graph related tasks were explored. A line of works propose strategies for scene graph generation from images~\cite{xu2017scenegraph,newell2017pixels,herzig2018mapping}. Johnson \etal~\cite{johnson2018image} introduced the reverse task of image generation from scene graphs, using a 2D layout as an intermediate representation between graphs and images, where layouts are decoded to images using a Cascade Refinement Network (CRN)~\cite{chen2017photographic} architecture. Later, a similar architecture was explored for image generation in an interactive form~\cite{ashual2019specifying} as well as for semantic image manipulation~\cite{dhamo2020semantic}. Herzig \etal~\cite{herzig2019canonical} proposed a model that uses canonical scene graphs, to improve robustness in terms of graph size and noise.   
Recently, Garg \etal \cite{Garg_2021_ICCV} generate scene graphs unconditionally and later synthesize images from the resulting graphs. 
Other related works explore image generation directly from layouts~\cite{zhao2018image,Sun_2019_ICCV,Sylvain2021ObjectCentricIG} or investigate 3D scene graphs~\cite{3DSSG2020,dhamo2021graph}.

\paragraph{Meta-learning}
Meta-learning or learning to learn was initially introduced for the few-shot classification problem. Model-agnostic Meta-Learning \cite{finn2017model} (MAML) is one of the most well-known works for few-shot image classification. MAML aims to optimize a model on a set of tasks using second-order gradient computation to obtain a model that adapts fast to newly seen tasks. Due to the extensive computation demands of MAML, a first-order approximation of it was proposed in Reptile~\cite{nichol2018first} with similar performance. These approaches have been mainly adopted in image classification and segmentation problems. A combination of meta-learning with GANs was introduced in \cite{zhang2018metagan} which employs adversarial training for few-shot image classification. However, the employment of meta-learning for the few-shot image generation task has been rarely explored. FIGR \cite{clouatre2019figr} uses meta-learning for few-shot image generation for small-scale datasets of black and white images, while few-shot image to image translation~\cite{liu2019few} generates images by adapting input from a source domain to a target domain. Despite the previous use of meta-learning for image generation, it has been only used on data with limited diversity and low resolution. To our knowledge, this is the first work on the few-shot generation of high-resolution scenes in the wild.

\section{Methodology}
\subsection{Problem Definition}

A task in meta-learning for few-shot image classification is defined as a set of image, label pairs. In our work, we define the task as scene graph and image pairs.
Given a set $\mathcal{D}$ of image $\mathcal{I}$ and scene graph $\mathcal{G}$ pairs, we define our initial dataset $\mathcal{D} = \{\mathcal{I}, \mathcal{G}\}$. The dataset is divided into different tasks based on the predefined task definition. In each iteration of the training phase, a task is randomly sampled and the scene graph to image model parameters are optimized on the selected task. In the test phase, the trained parameters are then used to fine-tune the model on specific target tasks. The main components of our method which are the image generation and meta-learning are described over the following sections.
\subsection{Image generation}
To tackle the task of scene graph to image generation we build upon SG2Im~\cite{johnson2018image} architecture as a foundation. SG2Im takes as input a scene graph, where the nodes correspond to the objects and the edges define relationships between them. The scene graph is processed by graph convolutional network (GCN), which operates on triplets subject-predicate-object, to propagate information along the edges. This results in processed per-node features, where each object has its own embedding vector that encodes information about itself as well as relationships with connecting objects. These embeddings are then used to predict a set of bounding boxes and segmentation masks for each object. The predicted boxes and masks are combined to project the GCN features to image space and obtain a scene layout. The next step in this pipeline is an image generator that receives a scene layout and produces an image corresponding to the given semantic definition. In order to force the network to produce both realistically looking and semantically correct images, two image discriminators are used on top of the generated image. The first one discriminates individual objects in a local context, while the second one classifies a picture as a whole.
We use different loss terms for training the model. Most of the losses are adopted from~\cite{johnson2015image}, but some extra loss terms are also used which we mention below. To prevent the model from generating trivial solutions, we employ the perceptual loss~\cite{johnson2016perceptual} $\lambda_p \mathcal{L}_p$ using the VGG network. There are two GAN losses defined, one for the whole image and one to make individual objects look realistic. These are defined as $\mathcal{L}_\text{GAN,global}$ and $\mathcal{L}_\text{GAN,obj}$ respectively. To ensure the quality of generated objects, the auxiliary classification loss  $\mathcal{L}_\text{aux,obj}$ is used. The loss for predicting the bounding boxes is defined by $\mathcal{L}_{box}$ which is calculated using the $\mathcal{L}_{1}$ loss between the predicted and ground truth bounding boxes. Finally, the image loss $\mathcal{L}_\text{im}$ which is the $\mathcal{L}_{1}$ distance between the predicted image and the ground truth image is used. \autoref{eq:taskloss} shows the task loss $\mathcal{L}_{\tau} $ definition.

\begin{equation}
  \begin{split}
      \mathcal{L}_{\tau} &=  \lambda_b %
       \mathcal{L}_{box} + \lambda_g \min_G \max_D \mathcal{L}_\text{GAN,global} \\ &\quad 
      + \lambda_o \min_G \max_D \mathcal{L}_\text{GAN,obj} + \lambda_a \mathcal{L}_\text{aux,obj} \\ &\quad 
      + \lambda_p \mathcal{L}_p + \lambda_\text{im} \mathcal{L}_\text{im},
  \end{split}
  \label{eq:taskloss}
\end{equation}
where $\lambda_b$,$\lambda_g$, $\lambda_o$, $\lambda_a$, $\lambda_p$, $\lambda_\text{im}$ are weighting values and
\begin{equation}
  \mathcal{L}_\text{GAN} = \mathop{\mathbb{E}}_{q\sim p_{\textrm{real}}} \log D(q) + \mathop{\mathbb{E}}_{q\sim p_{\textrm{fake}}} \log(1 - D(q)),
  \label{eq:ganloss}
\end{equation}
where $p_{real}$ refers to the real data distribution from the ground truth and $p_{fake}$ is the distribution of generated fake images or objects. The input to the discriminator is defined by $q$. 
\begin{figure*}[t]
    \centering
    \includegraphics[width=0.95\textwidth]{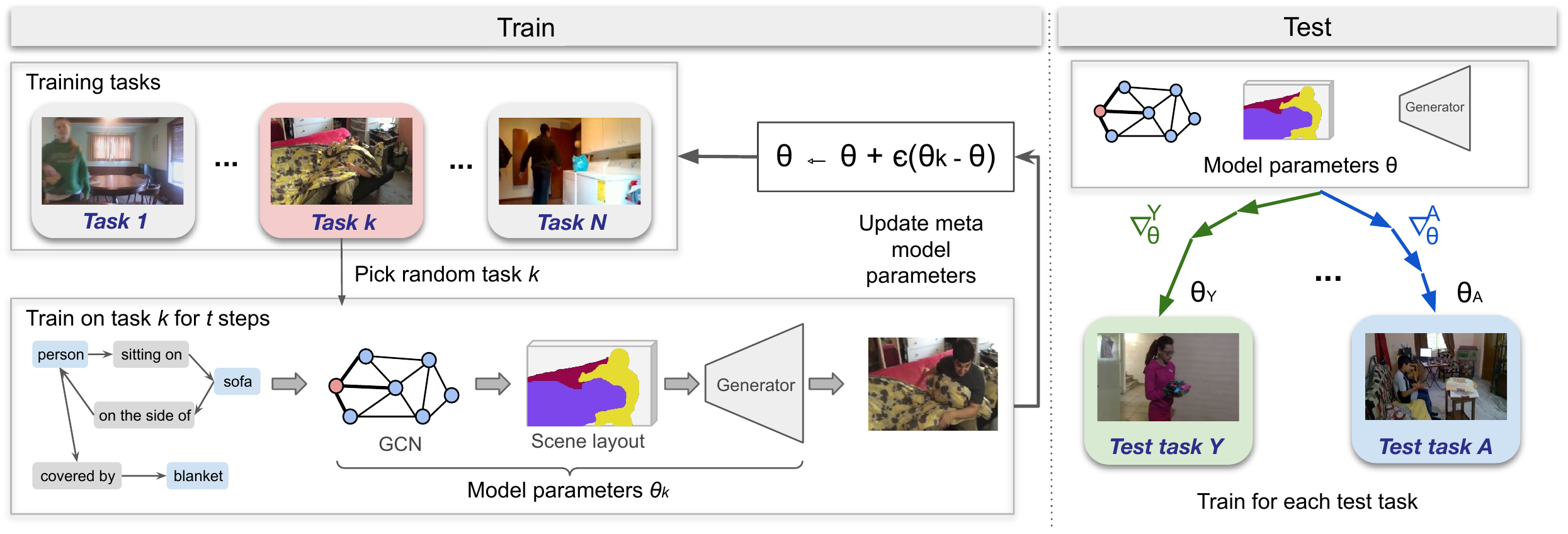} %
    \caption{\textbf{Method Overview}. Our method consists of two phases: Meta-training and Testing. In the meta-training phase, the model parameters $\theta$ are updated on a sampled task $l$ in each iteration. One task is a set of image and scene graph pairs that are unified in one group by some criteria. We pass a scene graph through a GCN to produce features from the nodes embeddings for creating a scene layout. The layout is passed to the generator, which synthesises the final image. In the testing phase, we fine-tune the model $\theta$ for a defined number of shots on each specific task which results in the generation of our final images.} %
    \label{fig:method}
\end{figure*}

In order to improve the image quality of the images generated by SG2Im and tackle a few-shot learning scenario, we adapt Reptile~\cite{nichol2018first} algorithm for GANs, similarly to how it is done in FIGR~\cite{clouatre2019figr}. However, FIGR is focused on the problem of unconditional image generation and experiments with images that have a single object drawn on them. We, on the other hand, have a setting conditioned on a scene graph with multiple objects, which is a more challenging problem compared to FIGR. We meta-train all components of the SG2Im pipeline on a diverse set of scenes and prove that such procedure enables high-quality image generation from the scene graph. 
\subsection{\nameMethod}
In this section, we introduce our method of Meta image generation from scene graphs (MIGS) and its components. Our primary goal is to be able to quickly and effectively adapt a model trained on a variety of images to a specific task with just a few training shots. To tackle this problem we refer to the meta-learning models~\cite{finn2017model,nichol2018first} and their application in a few-shot learning setting. To the best of our knowledge, we introduce a few-shot learning scenario for the scene graph to image generation for the first time. 

\paragraph{Task Definition}
Meta-learning models assume access to a set of tasks $T$, that consist of different learning problems $\tau$. Each of these tasks $\tau$ represent an image generation from scene graph problem and has its own set of image-graph pairs $\mathcal{D}_{\tau} = \{\mathcal{I}, \mathcal{G}\}_{\tau}$, that have been grouped together in one task based on certain criteria. Such criteria include the type of objects on the scene, specific background surroundings or any other attributes of either graphs, or images that unite them together. However, the splitting bases should be homogeneous across the whole set of tasks $T$. The task splitting criteria is dependent on the dataset characteristics. It can be defined based on the scene attributes such as time of the day, the context or by simply clustering the images into a set of clusters based on their visual attributes using an unsupervised clustering method such as \cite{van2020scan}.
\paragraph{Meta-learning}
A loss function on a task $\tau$ is denoted as $\mathcal{L}_{\tau}$. For simplicity we define $\mathcal{L}_{\tau}$ as a combination of all generator $\mathcal{L}_{\tau, G}$ and discriminator $\mathcal{L}_{\tau, D}$ losses in SG2Im model. Then our meta-learning goal is to find such initial model parameters $\theta$ that for a randomly selected task $\tau$ the loss $\mathcal{L}_{\tau}$ will be low after $k$ iterations with only a few data points available. In short, such objective is defined as
\begin{equation}
\min_{\theta}[\mathcal{L}_{\tau}\Big(U^{k}_{\tau}(\theta)\Big)], 
\end{equation}
where $U^{k}_{\tau}(\theta)$ denotes an operator that updates weights $\theta$ $k$ times using image-graph pairs from  $\mathcal{D}_{\tau}$.

In order to find such parameters $\theta$, we train the models with the Reptile algorithm~\cite{nichol2018first}. It comprises of inner and outer loops. In the inner loop $k$ iterations of operator $U$ are performed on the locally copied weights $\theta_l$  for a randomly sampled training task $l$. In an outer loop the weights vector of the meta-model $\theta$ are updated leveraging the difference between $\theta$ and $\theta_l$ computed in the inner loop. This updated can be summarized as: 
\begin{equation}
\theta \leftarrow \theta +
  \beta  \frac{1}{L} \sum_{l=1}^{L}  {(\theta_l-\theta)},
\end{equation}
where $L$ is number of tasks and $\beta$ is the meta learning rate. We perform such updates separately for the image generator model and two discriminators.

\paragraph{Testing}
For testing the trained meta-model, we first fine-tune the trained weights $\theta$ on the training split of each specific test task $l$ to obtain the final weights $\theta_l$ %
for this task. Then, we generate images from the (unseen) test set scene graphs of each task.

To fairly compare our meta-models to baseline methods, we use transfer learning on each of the non-meta models' weights $\Phi$ and fine-tune them on each task to obtain the corresponding $\Phi_l$. Then for each task $l$ we evaluate the images generated by $\theta_l$ and $\Phi_l$.

\begin{figure*}[t]
    \centering
    \includegraphics[width=0.97\textwidth]{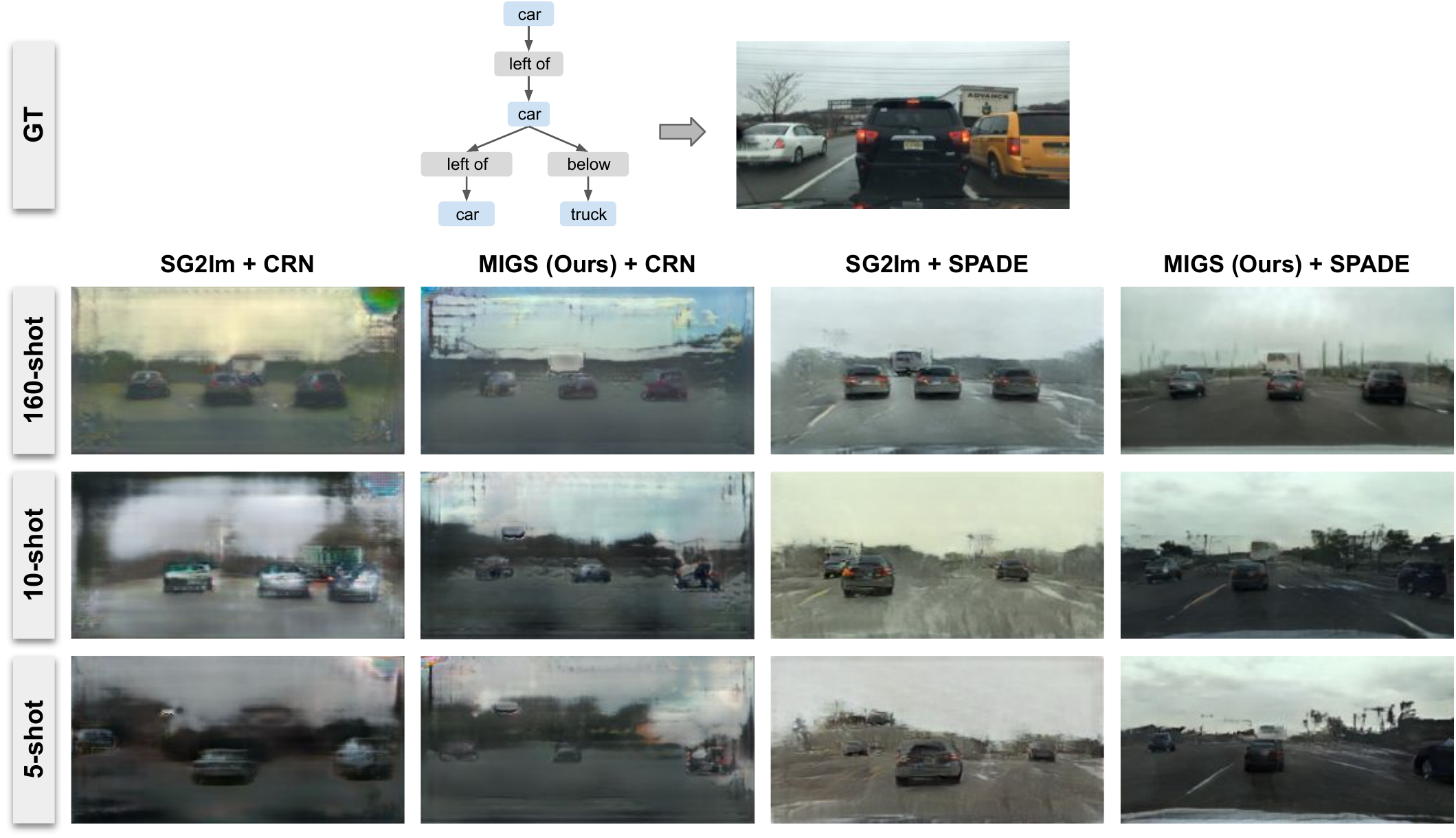} %
    \caption{Examples of images generated from BDD dataset for a task with the following attributes: day time: daytime; weather: rainy; driving scenario: highway. All images were generated from the scene graph defined on top. MIGS not only produces more realistic images but also has a higher accuracy when cross-referenced with the provided scene graph, i.e. MIGS + SPADE is the only model for which the truck is clearly seen in all three scenarios. }
    \label{fig:bdd_res}
\end{figure*}

\begin{figure*}[t]
    \centering
    \includegraphics[width=0.88\textwidth]{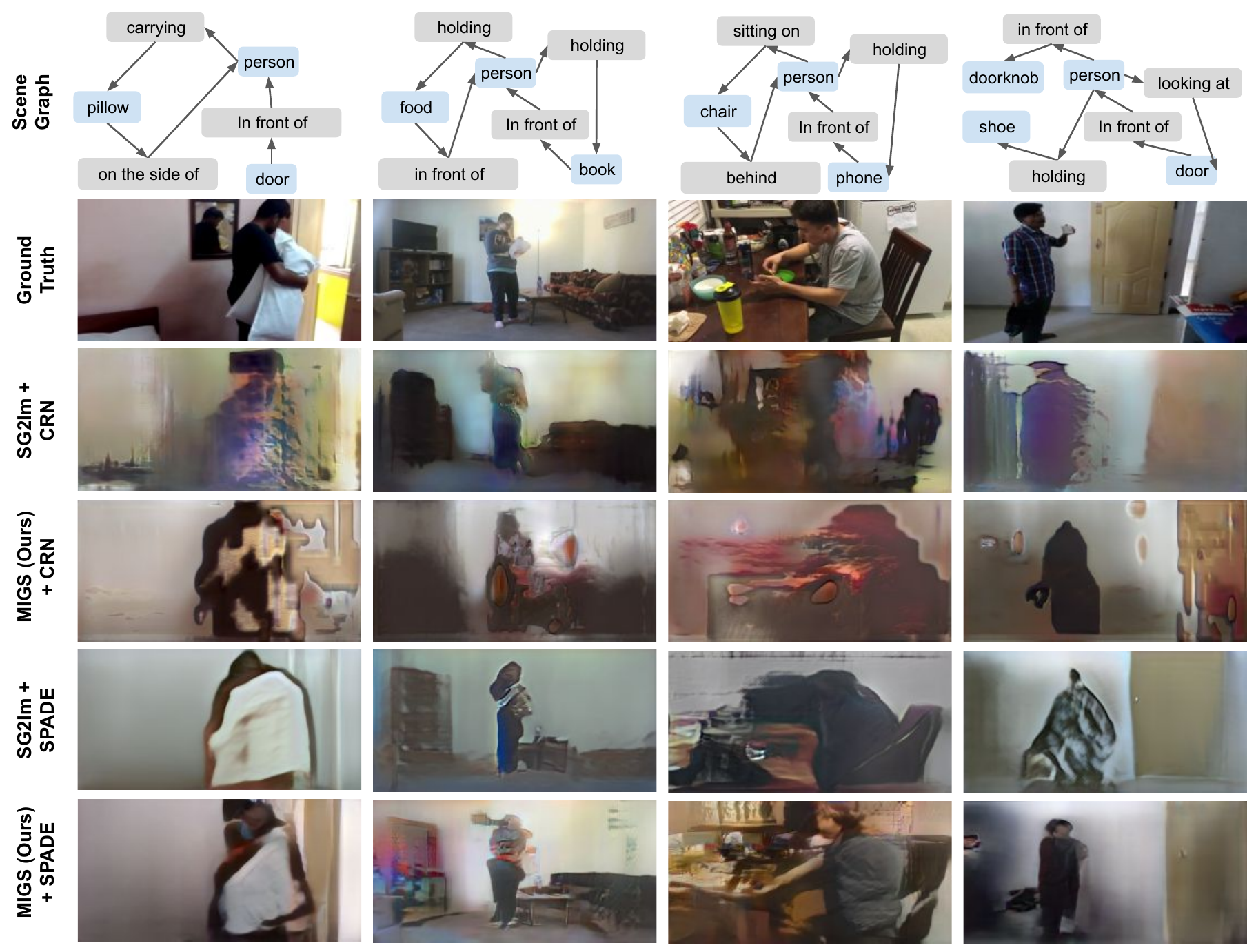} %
    \caption{Examples of images generated from the Action Genome dataset. Each column correspond to one task (i.e. a video sequence) that each model was fine-tuned on. The MIGS results on each video notably contain more details as opposed to the baseline counterpart.}
    \label{fig:ag_res}
\end{figure*}
\section{Experiments}
We evaluate our method on Berkeley Deep Drive~\cite{seita2018bdd100k} (BDD), Action Genome~\cite{ji2020action} (AG) and Visual Genome~\cite{krishna2017visual} (VG) datasets on different baselines. We show that our proposed method is independent of the generator architecture as the performance gain happens in two different generator architectures, namely CRN~\cite{chen2017photographic} and SPADE~\cite{park2019semantic}. To measure the quality and realism of the images generated by our proposed method quantitatively, Fr\'echet Inception Distance~\cite{heusel2017gans} (FID) and Kernel Inception Distance~\cite{binkowski2018demystifying} (KID) are reported. Moreover, generated image samples are compared to related work in diverse scenarios. We refer the reader to the supplement for the results with more metrics such as precision and recall~\cite{sajjadi2018precisionrecall} and the architecture details of the CRN, SPADE and GCN networks.

\subsection{Datasets}
As there are no established datasets for conditional image generation with meta-learning, we investigated and selected datasets %
that can be naturally categorized in a set of tasks. 

\paragraph{Berkeley Deep Drive~\cite{seita2018bdd100k} (BDD)} This dataset consists of images from city streets, residential areas, and highways in different scene conditions. As BDD does not contain scene graphs associated with images, we construct spatial scene graphs automatically, leveraging the ground truth bounding boxes, leading to six mutually exclusive spatial relationships: \texttt{left of}, \texttt{right of}, \texttt{above}, \texttt{below}, \texttt{inside}, and \texttt{surrounding}. As we are mostly interested in the objects and their relationships, we pre-process images and crop out the area of interest, which contains all the objects and as little background as possible.

For our meta-learning purpose, we split BDD into tasks using provided attributes for all images, such as time of day, weather conditions, and driving scenarios. We then filter out tasks that have less than 500 images available, which results in a total of 23 different meta-learning tasks. We use 20 tasks for training and validation and 3 for testing.

\paragraph{Action Genome~\cite{ji2020action} (AG)} The Action Genome dataset was originally designed for the action recognition problem. It consists of video frames of humans interacting with objects in a scene. We use human-object relationships labels from Action Genome directly to construct semantically meaningful scene graphs. 

As the AG dataset consists of a large number of videos with different actions, we use those videos as task splitting criteria. We remove the frames of all videos that do not have persons or only persons and no other objects, as it is impossible to construct meaningful scene graphs from them. Furthermore, we extract as tasks only the horizontal videos with at least 30 frames annotated by a human. This procedure provides us with 735 tasks which we split into 662 train and validation and 73 test tasks accordingly. 

\paragraph{Visual Genome~\cite{krishna2017visual} (VG)} The Visual Genome dataset provides images with their corresponding scene graphs and bounding box annotations which makes it suitable for the image generation from scene graphs problem. Since the VG dataset does not have specific annotations for scene attributes, we construct the tasks by splitting all the images into 100 classes using SCAN \cite{van2020scan} pre-trained on the imagenet dataset in an unsupervised manner. We use the first 60 clusters for the pre-training step and the last 40 clusters for evaluation.

\subsection{Experimental setup}
We train our models to generate images of size $128 \times 256$ for BDD and AG, and $64 \times 64$ for VG. For all meta-learning models, we use an inner learning rate of $0.0001$ and train for $10$ inner iterations with Adam optimizer. The outer loop has a learning rate of $1$ and uses SGD. The number of training iterations is dependant on the model and dataset, e.g. AG is more diverse than BDD and takes longer to converge. Meta-learning models on AG are trained for $40000$ outer loop iterations and models on BDD are trained for $30000$.

The baseline model, SG2Im \cite{johnson2018image} is trained on the same data as MIGS (all training tasks) in the pre-training step. During the testing phase, the pre-trained model is used as an initialization and is fine-tuned and tested in the few-shot setting similarly to MIGS. In all datasets, both models are trained until convergence. The SG2Im model is trained for 250k iterations on AG, 200k iterations on BDD, and 40k iterations on VG. The MIGS model is trained for 40k and 30k iterations on AG and BDD respectively, and 8k iterations on VG.

\subsection{Results}
The quantitative results of our experiments are shown in ~\autoref{tab:results_comp1}, \autoref{tab:results_comp2}. We compare our method to two baselines which are SG2Im with the original CRN decoder and another version with the SPADE network as the decoder to improve the generation quality. More qualitative results are provided in the supplementary material.

\paragraph{BDD Results} We show the performance of the mentioned models on BDD with different shot values ranging from 5 to 160 both quantitatively in ~\autoref{tab:results_comp1} and qualitatively in ~\autoref{fig:bdd_res}. Among all three experiments, we observe that FID and KID improve almost twice compared to the corresponding model with no meta-learning. We also perform a user study on BDD images, asking users to rank the quality of images from the same scene graph generated by different methods, and determine whether the scene represents the specified attribute. The results of the user study show that MIGS + SPADE is generally chosen as the most realistic method, MIGS + CRN as second and SG2Im + SPADE, SG2Im + CRN as third and fourth rank respectively. The exact percentages of rankings are shown in \autoref{tab:results_userstudy}. The values show the percentages of users choosing the method as the specified rank based on the image quality. For the attribute representation, MIGS + SPADE and MIGS + CRN stand as 1st and 2nd rank, while SG2Im + SPADE and SG2Im + CRN are ranked 3rd and 4th.

~\autoref{fig:bdd_res} shows example scene graph and images generated with baselines and our method on a single testing task from BDD. It is clear from these images, that our method outperforms all baselines and is able to generate realistic-looking images with a high level of detail even in an extremely challenging scenario where only 5 frames are available for training. Additionally, example images generated for a diverse set of training tasks may be seen on~\autoref{fig:teaser}. Our method successfully captures the differences in the scenes associated with daytime and driving scenario change.

\begin{table}[tbp]
    \centering
    \resizebox{0.85\columnwidth}{!}{
    \begin{tabular}{l|c|c|c|c|c|c|c}
    \toprule
        \multirow{2}{*}{Method} & \multirow{2}{*}{Decoder} & FID $\downarrow$ & KID $\cdot 10^3$ $\downarrow$ & FID $\downarrow$ & KID $\cdot 10^3$ $\downarrow$ & FID $\downarrow$ & KID $\cdot 10^3$ $\downarrow$\\ \cline{3-8}
         & & \multicolumn{2}{c|}{160-shot} & \multicolumn{2}{c|}{10-shot} & \multicolumn{2}{c}{5-shot}
        \\\midrule 
        SG2Im~\cite{johnson2018image} & CRN & 194 & 210 & 176 & 186.5  & 196.8 & 224.2 \\ %
        \nameMethod (Ours) & CRN & 158.5 & 156.4 & 157 & 158.4  & 183.5 & 187.6\\ %
        SG2Im & SPADE & 66.1 & 42.2  & 70.6 & 48.3 & 95.2 & 73.1 \\ %
        \nameMethod (Ours) & SPADE & \textbf{49.5} & \textbf{26.7}  &  \textbf{46.1} & \textbf{24}  &  \textbf{53.5} & \textbf{30.7}\\ %
        \bottomrule
    \end{tabular} }
    \caption{Quantitative results on BDD100k fine-tuned on 5,10 and 160 shots.}
    \label{tab:results_comp1}
\end{table}

\paragraph{AG Results}
We train the Action Genome model on all training images for each testing task (approximately 30 frames) and evaluate the model on the frames extracted from the full videos that were not used for training. The testing set has approximately 65000 images.

AG dataset is extremely challenging for the scene graph to image generation, as it contains labeling only for a few chosen objects on the image and most commonly those objects are quite small, e.g. phone or book. \autoref{fig:ag_res} shows the example images generated by our model as well as baselines on AG, while the quantitative performance on AG is shown in~\autoref{tab:results_comp2}. Due to the difficulty of the used dataset, it is quite expected that the results look different from the ones on BDD. However, even in this scenario, the number of details increases with our method compared to a corresponding baseline. This improvement can be also verified quantitatively from the substantial decrease in both FID and KID.

\begin{table}[htbp]
\noindent\parbox[t]{0.47\columnwidth}{
    \centering
    \resizebox{0.45\columnwidth}{!}{
    \begin{tabular}[t]{l|c|c|c}
    \toprule
        Method & Decoder & FID $\downarrow$ & KID $\cdot 10^3$ $\downarrow$\\ \midrule
        SG2Im~\cite{johnson2018image} & CRN & 198 & 163.4\\ %
        \nameMethod (Ours) & CRN & 174.5 & 137.8\\ %
         SG2Im & SPADE & 141.3 & 76.3  \\ %
        \nameMethod (Ours) & SPADE & \textbf{98.1} & \textbf{47.4} \\\bottomrule
        
    \end{tabular} }
    \caption{Quantitative results on Action Genome dataset compared to related work. }
    \label{tab:results_comp2}
    }
\hfill
\parbox[t]{0.49\columnwidth}{
\centering
 \resizebox{0.50\columnwidth}{!}{
\begin{tabular}[t]{l|c|c|c|c|c}
 \toprule
 \multirow{2}{*}{Method} & \multirow{2}{*}{Decoder} &\multicolumn{4}{c}{Rank}\\ \cline{3-6}
         &  & 1 (\%) & 2 (\%) & 3 (\%) & 4 (\%)\\ \midrule
        SG2Im~\cite{johnson2018image} & CRN & $15.79$ & $23.34$ & $26.55$ & $\textbf{35.58}$\\ %
        \nameMethod (Ours) & CRN & $24.46$ & $\textbf{25.57}$ & $25.16$ & $24.81$\\ %
         SG2Im & SPADE & $25.36$ & $24.60$ & $\textbf{28.29}$ & $21.74$ \\ %
        \nameMethod (Ours) & SPADE & $\textbf{34.72}$ & $26.48$ & $20.0$ & $17.79$
        \\\bottomrule
\end{tabular} }
\caption{User study ranking results on randomly sampled images from BDD dataset.} %
    \label{tab:results_userstudy}
}
\end{table}

\begin{figure*}[ht!]
    \centering
    \includegraphics[width=0.88\textwidth]{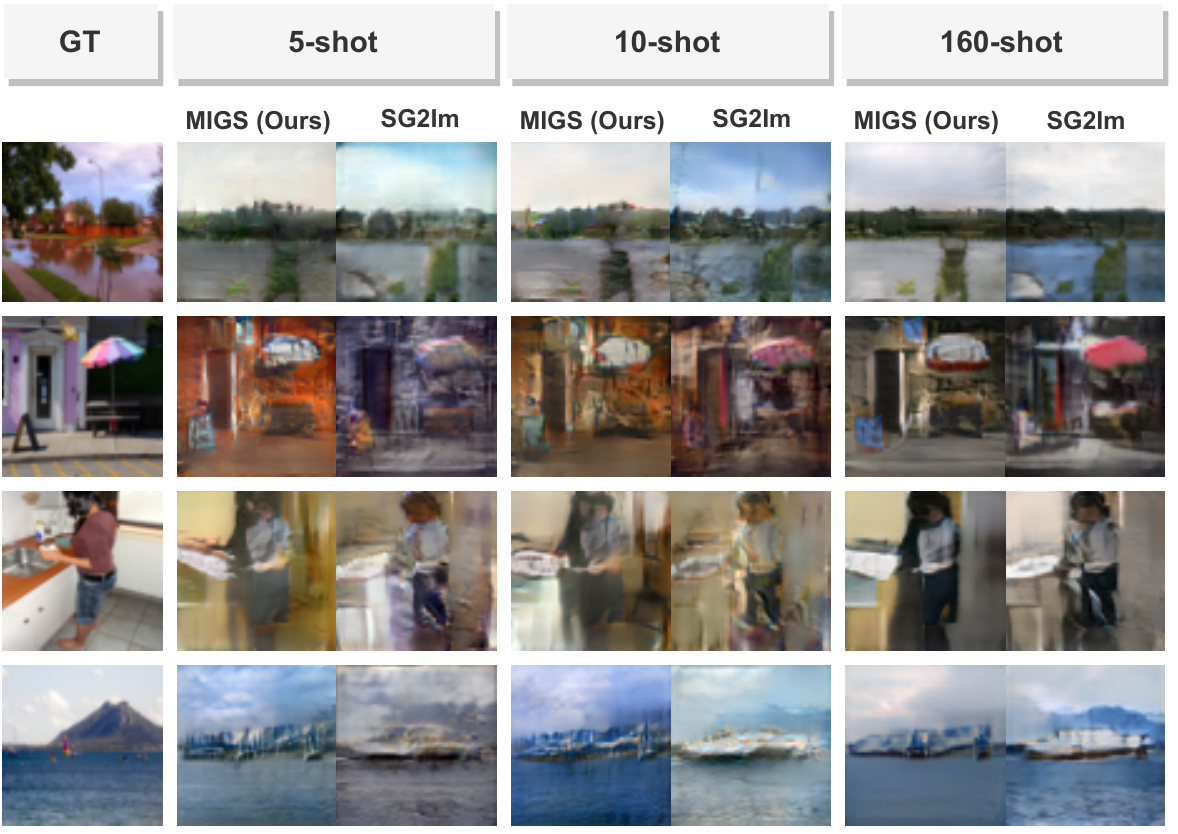}
    \caption{Example images generated with MIGS + SPADE and SG2Im + SPADE on the Visual Genome dataset.}
    \label{fig:vg_res}
\end{figure*}

\paragraph{VG Results}
The performance of MIGS compared to SG2Im~\cite{johnson2018image} on the VG dataset is presented in \autoref{tab:results_comp_vg} and \autoref{fig:vg_res}. As shown in the results, MIGS outperforms the baseline in all metrics for different shot values, even with less number of training epochs. Despite the higher diversity of the images in the VG dataset and their wild nature, MIGS is able to generate images which look more realistic compared to the baseline.

\begin{table}[tbp]
    \centering
    \resizebox{0.9\columnwidth}{!}{
    \begin{tabular}{l|c|c|c|c|c|c|c}
    \toprule
        \multirow{2}{*}{Method} & \multirow{2}{*}{Decoder} & FID $\downarrow$ & KID $\cdot 10^3$ $\downarrow$ & FID $\downarrow$ & KID $\cdot 10^3$ $\downarrow$ & FID $\downarrow$ & KID $\cdot 10^3$ $\downarrow$\\ \cline{3-8}
         & & \multicolumn{2}{c|}{160-shot} & \multicolumn{2}{c|}{10-shot} & \multicolumn{2}{c}{5-shot}
        \\\midrule 
        SG2Im \cite{johnson2018image} (All epochs) & SPADE & 55.20 & 35.54  & 81.42 & 59.39 & 91.79 & 68.52 \\ %
        \nameMethod (Ours, 1/3 epochs) & SPADE & 54.83 & 34.21 &  76.56 & 52.02  &  84.87 & 59.38 \\ %
        \nameMethod (Ours, All epochs) & SPADE & \textbf{54.24} & \textbf{29.00} &  \textbf{75.96} & \textbf{50.69}  &  \textbf{83.54} & \textbf{55.28} \\ %
        \bottomrule
    \end{tabular}}
    \caption{Quantitative results on VG fine-tuned on 5, 10 and 160 shots.}
    \label{tab:results_comp_vg}
\end{table}

\subsection{Discussion}
Our experiments show that using meta-learning for the task of image generation from scene graphs, outperforms the respective baselines almost twice, in terms of the employed metrics. %
The proposed method is shown to be advantageous in all scenarios even when using only 5 training samples. %
The results obtained on BDD qualitatively differ, which can be attributed not only to a more challenging setting of the AG but also to the unsuitability of this dataset for the image generation task. As it is intended for action recognition on video, only the objects that the person interacts with are annotated. It is thus common for some instances of the objects to be ignored in certain frames, until the action takes place, or be unannotated through the whole video if not used by the main actor. %
Such discrepancies may confuse the model and result in poor quality. Additionally, this dataset contains a lot of very small objects compared to the frame size. We believe that with a better dataset for semantically meaningful scene graphs the model should demonstrate results similarly to BDD. The results on the VG dataset show that using a simple yet effective task construction scheme, such as clustering, which could be applied to any other dataset, combined with the meta-learning approach can improve the performance of the image generation. The effect of task construction on the performance of meta-learning is an interesting topic and leaves room for research for future.

\section{Conclusion}
We propose \nameMethod, a meta-learning approach for the generation of images from scene graphs. To our knowledge, this is the first work for few-shot image generation of scenes in the wild. The proposed method could be applied to a different range of generator architectures and different datasets. We plan to replace the current SG2Im framework with \cite{herzig2019canonical} as part of the future work. The results of the evaluation on three datasets show that our proposed meta-learning-based image generation scheme proves to improve the quality of generated images significantly in all scenarios. We show that it is possible to generate high-quality images only given 5 shots of data. The performance gain compared to previous work is shown both quantitatively and qualitatively in the results.
\paragraph*{Acknowledgement}
We gratefully acknowledge the Munich Center for Machine Learning (MCML) with funding from the Bundesministerium für Bildung und Forschung (BMBF) under the project 01IS18036B. We are also thankful to Deutsche Forschungsgemeinschaft (DFG) for supporting this work, under the project 381855581.

\bibliography{ref}

\end{document}


\def\httilde{\mbox{\tt\raisebox{-.5ex}{\symbol{126}}}}
\newcommand{\nameMethod}{MIGS}

\def\thefootnote{*}\footnotetext{The first two authors contributed equally to this work.}

\addauthor{Azade Farshad\thefootnote{}}{azade.farshad@tum.de}{1} %
\addauthor{Sabrina Musatian\thefootnote{}}{sabrina.musatian@tum.de}{1}
\addauthor{Helisa Dhamo}{helisa.dhamo@tum.de}{1}
\addauthor{Nassir Navab}{nassir.navab@tum.de}{1}

\addinstitution{
Technical University of Munich\\
 Munich, Germany
}
\title{Supplementary Material: MIGS: Meta Image Generation from Scene Graphs}
\runninghead{MIGS}{Meta Image Generation from Scene Graphs}
\maketitle

\section{Experiment On Memorization}
\begin{figure*}
    \centering
    \includegraphics[width=\textwidth]{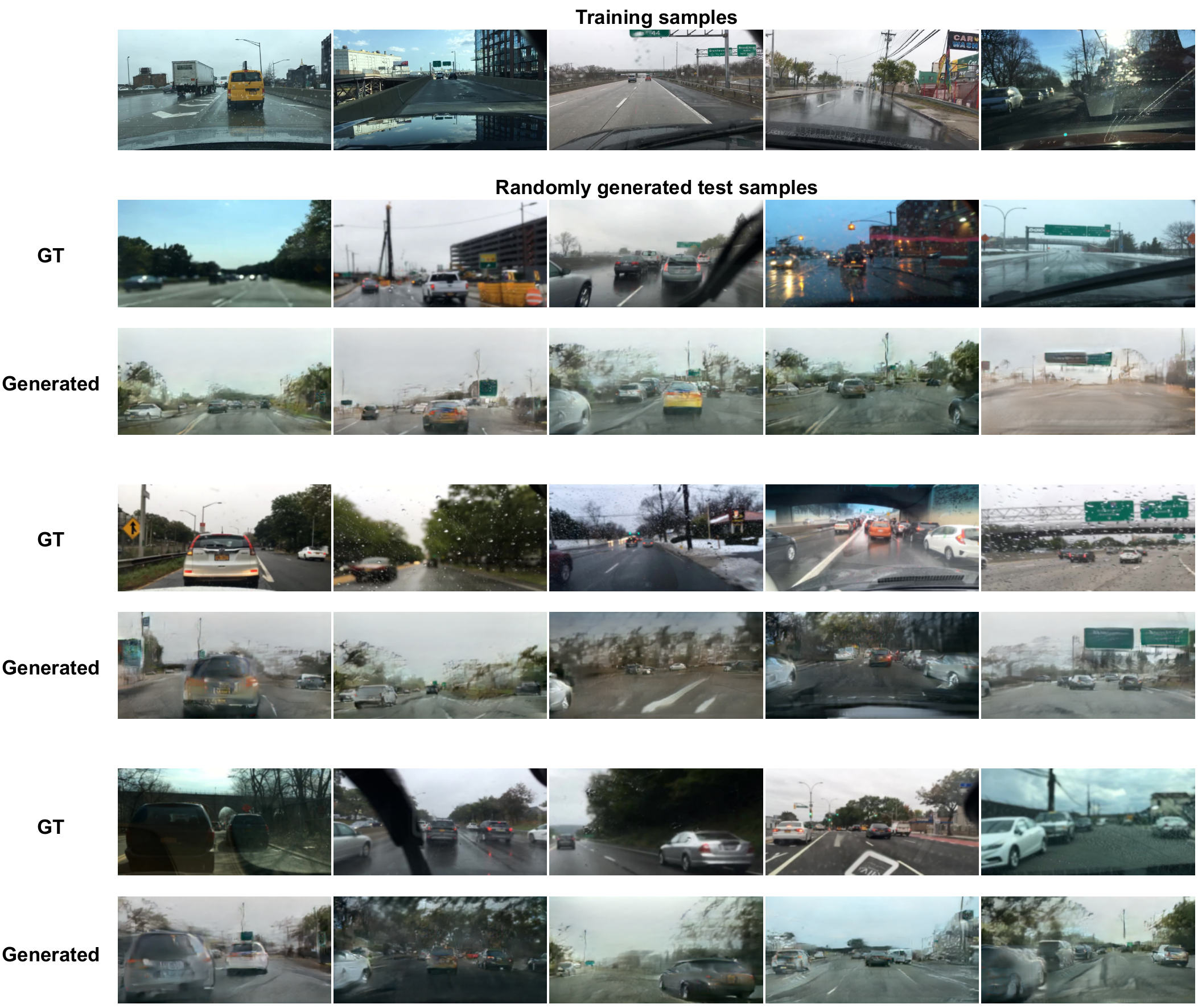}
    \caption{Illustration of 5-shot learning results on Berkeley Deep Drive (BDD) with MIGS + SPADE model for effect of memorization.}
    \label{fig:memorization}
\end{figure*}

In \autoref{fig:memorization} we show training samples for a 5-shot training task and generated images from the test set of the same task, along with the corresponding ground truth images. As it can be seen in the figure, the generated images are different from the 5 training samples in terms of e.g. car colors or shapes. It should be noted that the model has generated cars with different colours than the ones it was exposed to by the training samples. The diversity of the generated images is shown to be better than the original sg2im model given the Precision and Recall metric in \autoref{tab:results_comp_bdd_prec_rec}, and FID and KID metrics. In the testing phase, for each task consisting of different images and their corresponding scene graphs, 5 different images (in the case of 5-shot) are sampled and used for fine-tuning the model. Then other images (with different scene graphs) from the same task with similar attributes are sampled. Their scene graph is used to generate images and the generated images are compared to the ground truth ones.

\section{Additional Quantitative Results}
In this section we provide quantitative evaluation of our method compared to baseline using precision ($F_{1/8}$) and recall ($F_8$) metrics [30] for BDD (\autoref{tab:results_comp_bdd_prec_rec}) and AG(\autoref{tab:results_comp_ag_prec_rec}) datasets.
\begin{table}[tbp]
    \centering
    \resizebox{0.9\columnwidth}{!}{
    \begin{tabular}{l|c|c|c|c|c|c|c}
    \toprule
        \multirow{2}{*}{Method} & \multirow{2}{*}{Decoder} & F_8 $\uparrow$ & F_{1/8} $\uparrow$ & F_8 $\uparrow$ &  F_{1/8}  $\uparrow$ & F_8 $\uparrow$ &  F_{1/8}  $\uparrow$\\ \cline{3-8}
         & & \multicolumn{2}{c|}{160-shot} & \multicolumn{2}{c|}{10-shot} & \multicolumn{2}{c}{5-shot}
        \\\midrule 
        SG2Im& CRN & 0.101 & 0.135 & 0.063 & 0.131  & 0.06 & 0.057 \\ %
        \nameMethod (Ours) & CRN & 0.06 & 0.176 & 0.052 & 0.123  & 0.05 & 0.06\\ %
        SG2Im & SPADE & 0.486 & 0.612  & 0.462 & 0.45 & 0.329 & 0.438 \\ %
        \nameMethod (Ours) & SPADE & \textbf{0.7} & \textbf{0.86}  &  \textbf{0.79} & \textbf{0.854}  &  \textbf{0.74} & \textbf{0.823}\\ %
        \bottomrule
    \end{tabular} }
    \caption{Additional quantitative results on BDD100k fine-tuned on 5,10 and 160 shots.}
    \label{tab:results_comp_bdd_prec_rec}
\end{table}
On both datasets MIGS + SPADE outperforms the corresponding baseline by almost twice. Such supremacy means that the images generated with the meta-learning approach are more realistically looking (precision) and cover more modes of the underlying real data distribution (recall). The results obtained for MIGS + CRN are comparable to the ones of the baseline.

\begin{table}[tbp]
    \centering
    \resizebox{0.45\columnwidth}{!}{
    \begin{tabular}[t]{l|c|c|c}
    \toprule
        Method & Decoder &  F_8 $\uparrow$ & F_{1/8} $\uparrow$\\ \midrule
        SG2Im & CRN & 0.217 & 0.116\\ %
        \nameMethod (Ours) & CRN & 0.167 & 0.09\\ %
         SG2Im & SPADE & 0.59 & 0.31  \\ %
        \nameMethod (Ours) & SPADE & \textbf{0.6} & \textbf{0.5} \\\bottomrule
        
    \end{tabular} }
    \caption{Additional quantitative results on Action Genome dataset compared to related work. }
    \label{tab:results_comp_ag_prec_rec}
\end{table}

\section{Extra Qualitative Results}

\begin{figure*}
    \centering
    \includegraphics[width=\textwidth]{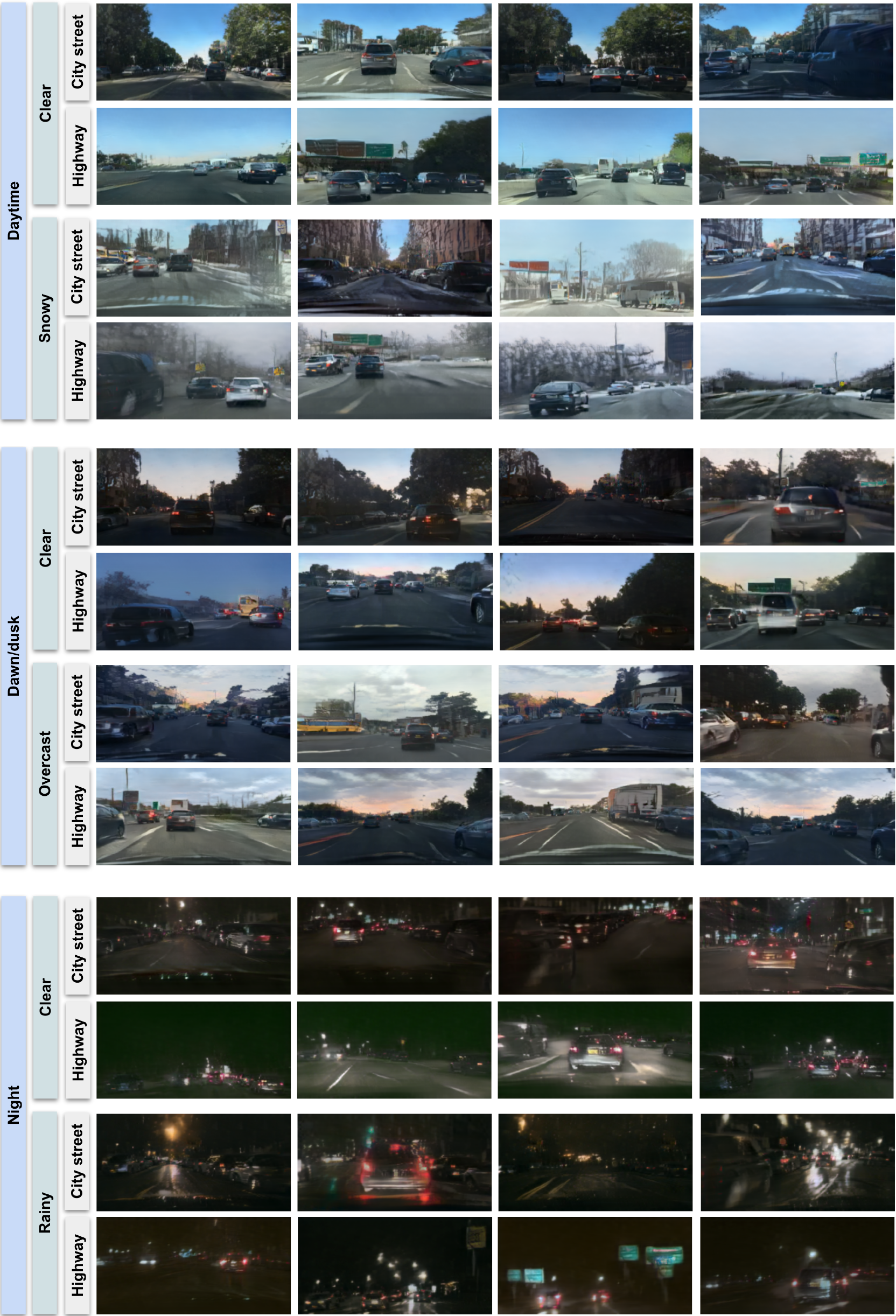}
    \caption{Illustration of 160-shot learning results on Berkeley Deep Drive (BDD) with MIGS + SPADE model.}
    \label{fig:160shot}
\end{figure*}

\begin{figure*}
    \centering
    \includegraphics[width=\textwidth]{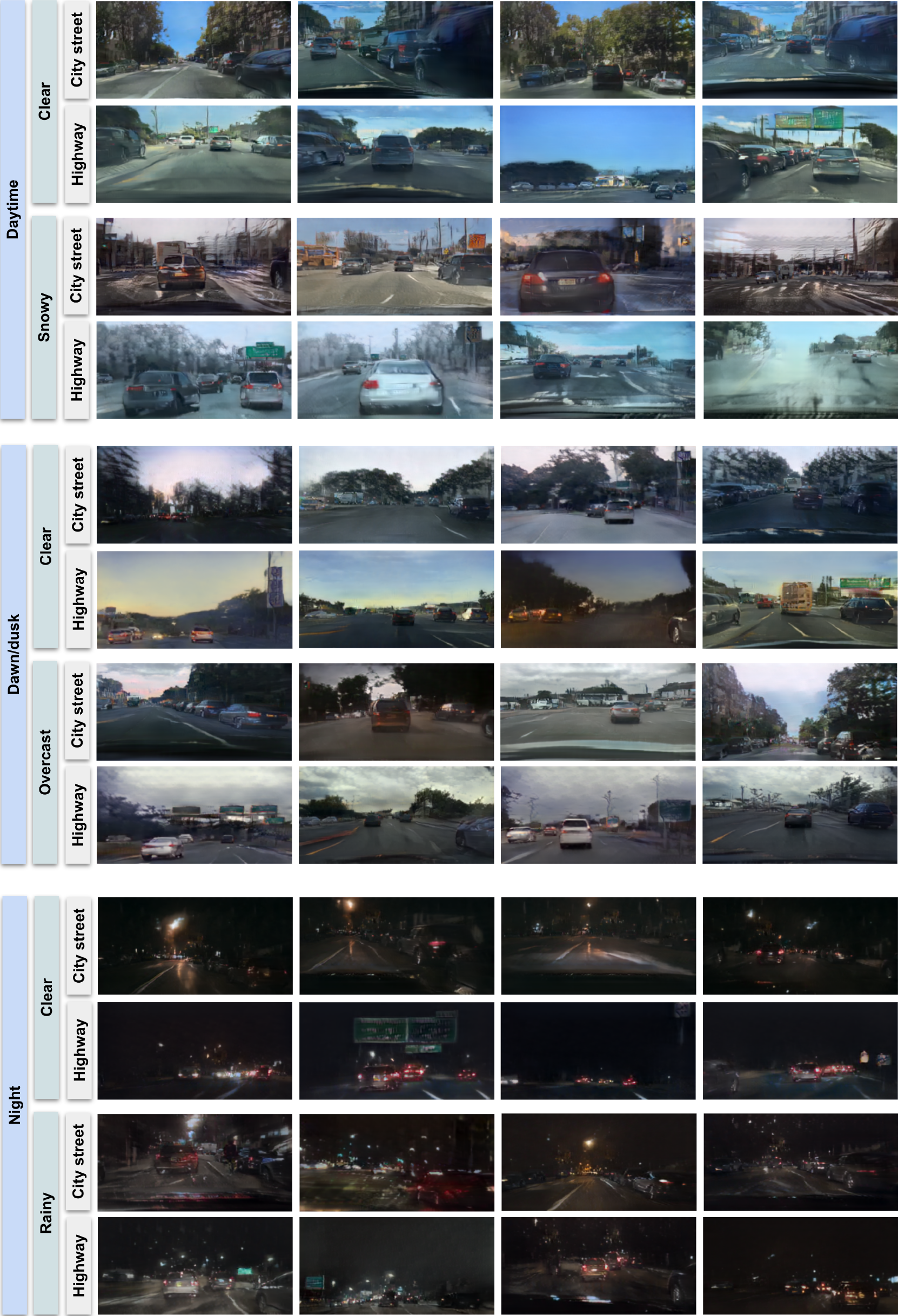}
    \caption{Illustration of 10-shot learning results on Berkeley Deep Drive (BDD) with MIGS + SPADE mo    del.}
    \label{fig:10shot}
\end{figure*}

\begin{figure*}
    \centering
    \includegraphics[width=\textwidth]{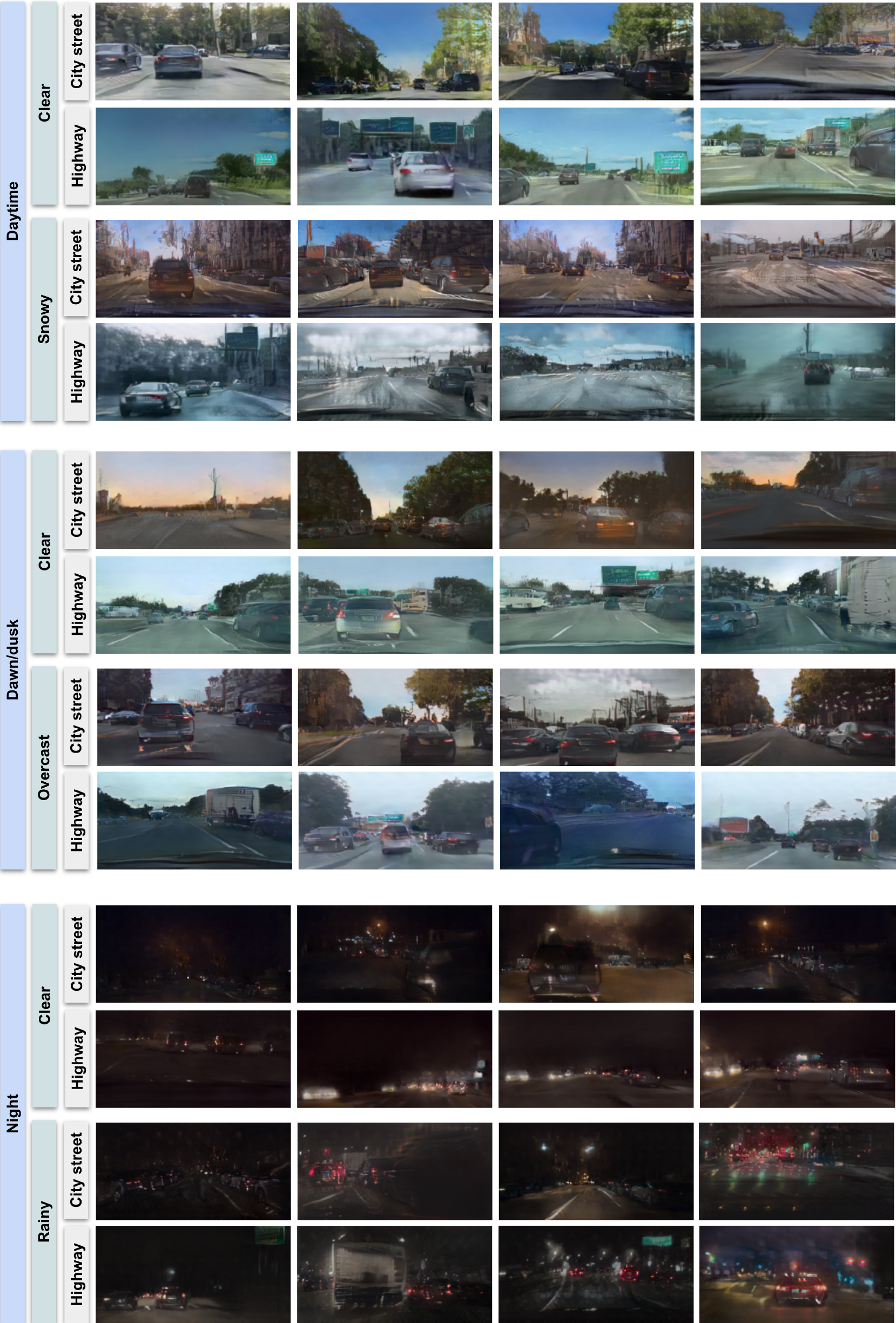}
    \caption{Illustration of 5-shot learning results on Berkeley Deep Drive (BDD) with MIGS + SPADE model.}
    \label{fig:5shot}
\end{figure*}

Here we demonstrate additional qualitative results of MIGS on the BDD dataset in 160-shot learning (Figure~\ref{fig:160shot}), 10-shot learning (Figure \ref{fig:10shot}) and 5-shot learning (Figure \ref{fig:5shot}). We observe that the model generates compelling results for a diverse set of tasks. All images were generated using MIGS with SPADE generator. Both 160 and 10-shot learning models are able to generate various sets of images even within one particular task. The results of these two models are comparable in quality and both are able to depict such fine-grained details as clouds in the overcast task or glares from the rain on the road. 5-shot learning can produce realistically looking images for many tasks, but struggles with reproducing significantly differently looking images within one particular task. It also cannot generate as many details as can be seen in 160 and 10-shot learning models. This behavior is not surprising, as the model sees only a very limited amount of training images, so it might be impossible to depict such diversity from this set.

\begin{figure}
    \centering
    \includegraphics[]{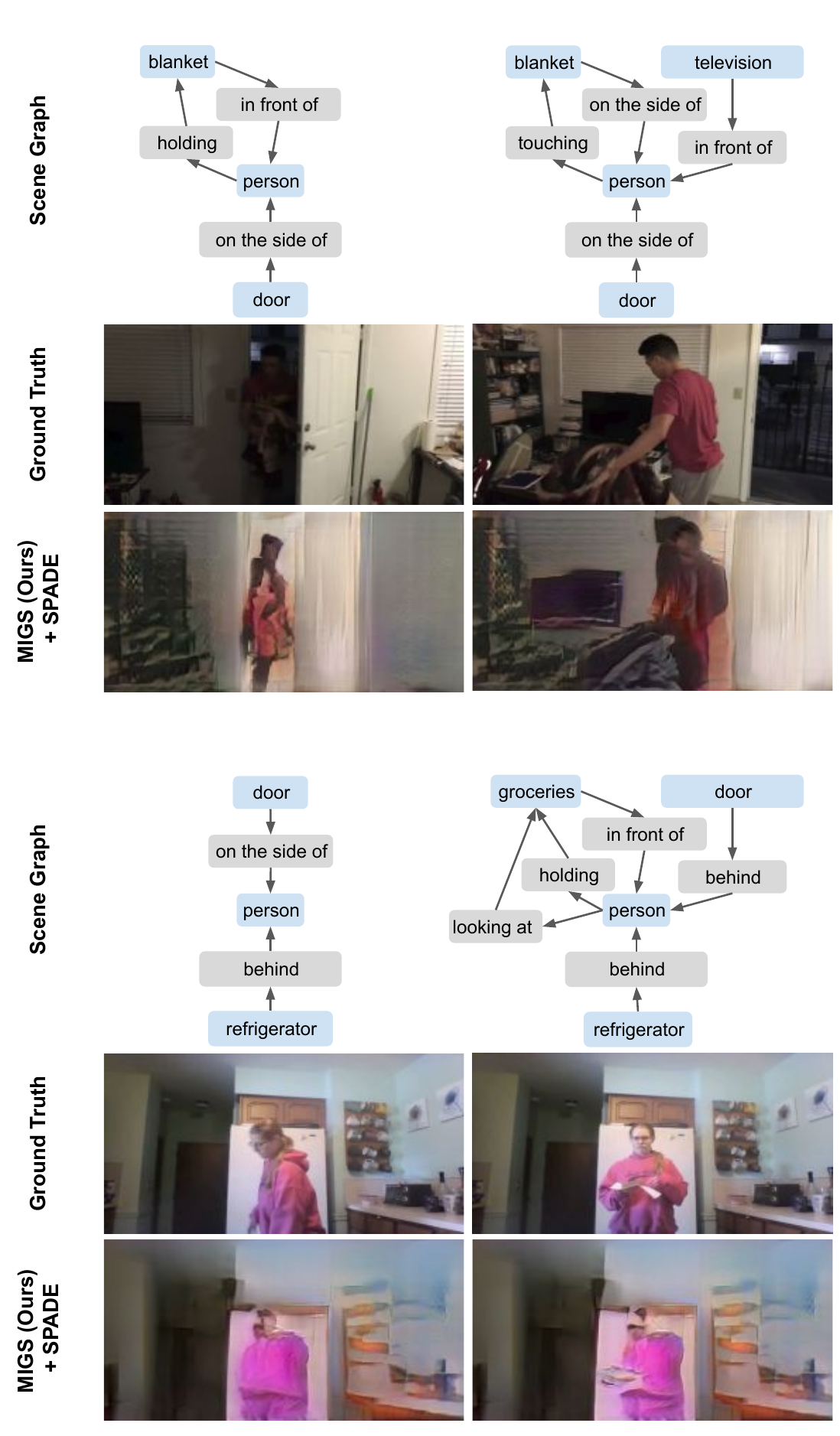}
    \caption{Additional examples of images generated with MIGS + SPADE trained on particular video classes from Action Genome.}
    \label{fig:ag_extra}
\end{figure}

Additionally, we provide more qualitative results on AG dataset (Figure~\ref{fig:ag_extra}) to demonstrate that our method can correctly capture the semantic relationships between the objects, specified by the scene graph in different scenarios.

\section{User study}
For the perceptual study, $600$ images were generated randomly for three different scene attributes (daytime, dawn/dusk, night) with $200$ examples of each. Each image was seen by $3$ workers. In each example, the user receives four images in random order -- representing our four methods in study -- and is asked to provide a ranking among them. In addition, we provide a checkbox for each image, through which the user can indicate whether a certain attribute is met. 

\section{Architecture details}

\paragraph{CRN} The CRN variant of the decoder network contains $5$ cascaded refinement blocks, which have namely 1024, 512, 256, 128 and 64 channels. 
Every block consists of two $3\times3$ convolutions, each followed by batch norm and leaky ReLU. The output of each module is concatenated with the initial input to the CRN, re-scaled to the feature resolution.

\paragraph{SPADE} The SPADE decoder consists of $5$ residual blocks, which have namely 1024, 512, 256, 128 and 64 channels. Instead of the semantic map in the original implementation, here we use the layout to modulate the layer activations in each block. The global discriminator $D_{global}$ contains two scales. 

\paragraph{GCN} The GCN network consists of $5$ layers. Each layer processes triplets of subject - predicate - object embeddings, which are obtained by feeding each semantic label in an embedding layer. Every layer consists of three steps. First, the propagation layer (a two-layer MLP) receives the concatenated triplet feature and results in a 128 channels output. Second, the aggregation layer computes the average of features that correspond to a certain node. Third, the update layer applies a final processing of each node feature via another two-layer MLP. Both MLPs above have a hidden layer of 512 channels. The input embeddings of the objects and predicates have $128$ dimensions each.
The last layer of the GCN returns the node features (128 channels), binary masks ($16\times16$) and bounding box prediction by applying a two-layer MLP with a hidden layer size of $128$.

